\documentclass[conference]{IEEEtran}
\IEEEoverridecommandlockouts
\usepackage{cite}
\usepackage{amsmath,amssymb,amsfonts}
\usepackage{algorithmic}
\usepackage{graphicx}
\usepackage{textcomp}
\usepackage{xcolor}
\usepackage{color,soul}
\usepackage[hyphens]{url}
\usepackage{multirow}
\usepackage{csquotes}
\usepackage{caption}
\usepackage{subcaption}
\usepackage{enumitem}
\usepackage{hyperref}
\def\BibTeX{{\rm B\kern-.05em{\sc i\kern-.025em b}\kern-.08em
    T\kern-.1667em\lower.7ex\hbox{E}\kern-.125emX}}

\title{With Shared Microexponents, \\A Little Shifting Goes a Long Way}

\author{\textbf{Bita Rouhani$^*$\thanks{*Corresponding Author. Email: \href{bita.rouhani@microsoft.com}{bita.rouhani@microsoft.com}}, Ritchie Zhao, Venmugil Elango, Rasoul Shafipour, Mathew Hall, Maral Mesmakhosroshahi}, 
\\ \textbf{Ankit More, Levi Melnick, Maximilian Golub, Girish Varatkar, Lei Shao, Gaurav Kolhe, Dimitry Melts}, \\ \textbf{Jasmine Klar, Renee L'Heureux, Matt Perry, Doug Burger, Eric Chung}\\
Microsoft \\ 
\textbf{Zhaoxia (Summer) Deng, Sam Naghshineh, Jongsoo Park, Maxim Naumov} \\
Meta
 }

\newtheorem{theorem}{Theorem}

\usepackage{etoolbox}
\makeatletter
\patchcmd{\@makecaption}
{\scshape}
{}
{}
{}
\makeatother
\begin{document}
\maketitle
\thispagestyle{plain}
\pagestyle{plain}

\begin{abstract}
This paper introduces Block Data Representations (BDR), a framework for
exploring and evaluating a wide spectrum of narrow-precision formats for deep learning. 
It enables comparison of popular quantization standards, and through BDR, 
new formats based on shared microexponents (MX) are identified, 
which outperform other state-of-the-art quantization
approaches, including narrow-precision floating-point and block floating-point. 
MX utilizes multiple levels of quantization scaling with ultra-fine
scaling factors based on shared microexponents in the hardware. The effectiveness of MX is demonstrated
on real-world models including large-scale generative pretraining and inferencing, and
production-scale recommendation systems.
\end{abstract}

\section{Introduction}\label{sec:intro}
The disruptive AI capabilities that are rapidly emerging are driven primarily by scale.  
With each order-of-magnitude increase, models (particularly large language models such as GPT-3) show amazing and surprising capabilities emerging. 
Unfortunately, that same scale results in high computation and energy costs to train and serve these models.  
 
The AI industry is applying many techniques to reduce the costs of models at each scale point.  
Quantization of tensor values is one important instance of these techniques, in which the individual tensor values are cast from FP32 to a cheaper numeric standard.  
Given the importance, there has been a marked increase in quantization formats, papers, and implementations over the past few years.  
Nvidia's support for an FP8 standard is one important recent example~\cite{nvidia2022fp8}.

Quantization standards aim to balance three factors that can trade off against one another:  

 \begin{itemize}[leftmargin=*]
   \item Efficiency, a combination of compute efficiency (energy and silicon area for dot products) and memory efficiency (the bits required to store or communicate a tensor).
   \item Accuracy (or ``model performance'' in AI parlance), the amount of loss (reduced quality of results) incurred by the quantization compared to FP32 precision.
   \item Friction, the user effort required to apply the quantization format effectively.  Friction can arise from uneven results across different model classes, or from the need to tweak software parameters to achieve the target accuracy.
 \end{itemize}

Many data formats have been proposed, and the design space of possible new ideas is rich.  
The diversity of data formats makes for a complex space given subtle trade-offs between efficiency, accuracy, and friction.  In this paper, we describe a framework that encompasses much of the quantization design space.  
This framework, which we call Block Data Representations (BDR) defines the ways to quantize and scale sets of tensor values collectively.  
Design points in this space may have large sets ($\sim 1K$) or small sets ($\sim 10$), they may use integer or floating-point scaling factors, the scaling factors may be applied by hardware or software, and there may be multiple levels of scaling among different sets.
 
Using the BDR framework, we survey and evaluate many of the popular quantization standards, including INT4, INT8, variants of FP8~\cite{nvidia2022fp8}, VSQ~\cite{dai2021vs}, and block floating-point variants (including MSFP~\cite{darvish2020pushing}).  To evaluate quantization accuracy, we define a statistical methodology that computes quantization signal-to-noise ratio, validating it against a range of real models.  
To measure hardware efficiency, we define a dot product engine in a leading process node that enables the computation of synthesized power and area for different quantization configurations.  
Leveraging these two models, we sweep hundreds of configurations in the design space and construct a Pareto frontier of the best accuracy and efficiency quantization design points.
 
\begin{figure*}[ht]
   \centering
   \includegraphics[width=0.9\textwidth]{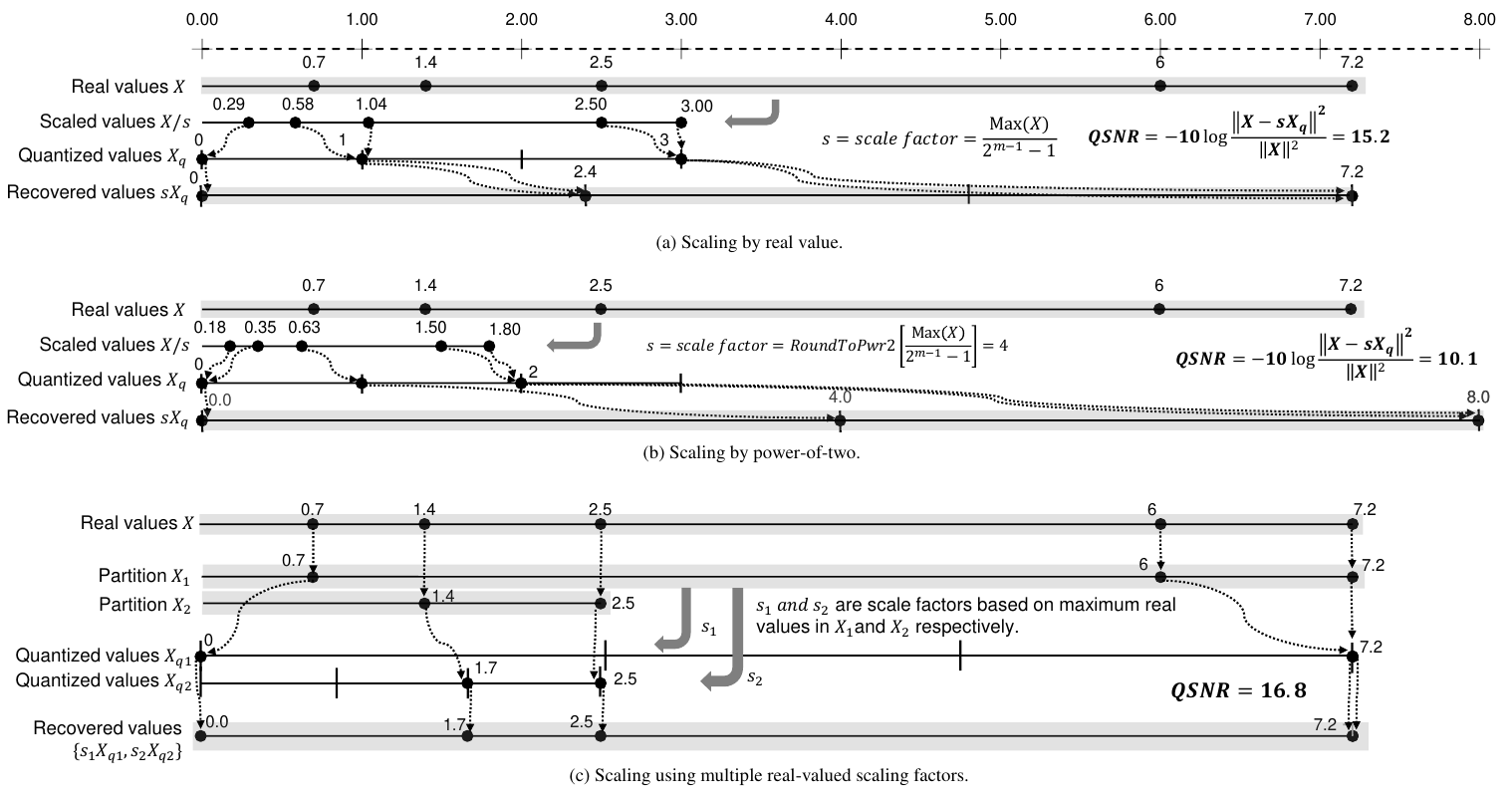}
    \caption{(a) INT-based quantization with FP32-based scaling based on maximum value,
    (b) Scaling is constrained to powers of two for hardware efficiency,
    (c) Multiple scaling factors enable increased resolution and reduced quantization error.} 
    \label{figure:scale-ex}
    \vspace{-1em}
\end{figure*}

With the BDR framework, we further describe a new numerical approach to quantization formats based on shared microexponents (MX).  MX leverages multiple levels of scaling, but unlike prior multi-level approaches such as FP8 and VSQ, it uses fine-grained scaling factors ($\sim 10$ elements) and ultra-fine-grained level-two scaling factors (one bit exponent shared by two elements), both set in the hardware.  
The fine-grained scaling factor reduces the numerical blast radius from outliers, and the ultra-fine-grained second-level scaling factor provides additional noise reduction, with no software friction since all scaling factors are automatically set by hardware.
 
The accuracy and efficiency models demonstrate that the MX approach creates a new Pareto frontier which makes many of its design points best in class compared to the other popular quantization approaches.  
We identify three members of the MX family ($4$, $6$, and $9$ bits, respectively) as being the most desirable to include as a set in silicon implementations.  
These three design points sit close to the new Pareto frontier.  
They provide a range of accuracy/efficiency/friction design points that align with the scenarios most commonly used by practitioners in both training and inference.  
Finally, they use a relatively small number of elements for the shared level-one scaling factor, making them easy to incorporate in a wide range of tensor unit designs and more amenable to fine-grained sparsity support than larger block sizes.
\section{Quantization Overview}
\label{sec:quantintro}

\begin{figure*}[ht]
   \centering
   \includegraphics[width=6.8in]{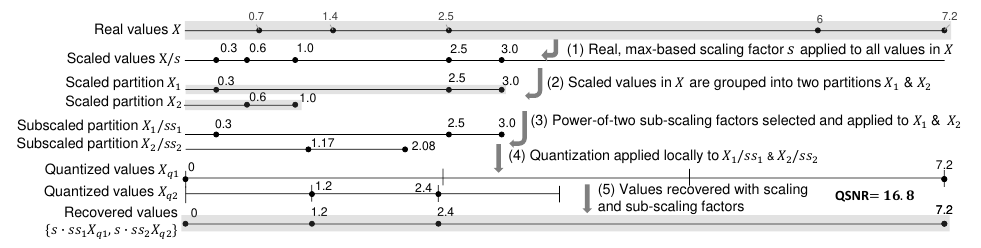}
    \caption{Example of a two-level scaling approach approximating ideal FP32-based scaling on two partitions. Low-cost sub-scale factors encoded in powers of two are composed with a top-level FP32 scaling factor to finetune scaling within each partition.}
    \label{figure:twolevelex}
    \vspace{-1em}
\end{figure*}

In deep learning, quantization refers to the process of 
numerically constraining a model's set of weights, activations, and gradients
to compact representations, e.g., from 32-bit floating-point values to 4-bit INTs.
Quantization can be used to improve the efficiency of memory and communication, or to 
accelerate computations during both training and inferencing.

\vspace{4pt}\noindent\textbf{INT Quantization.}
Uniform INT-based quantization is a popular method that maps real
values to INTs.  As illustrated in Figure~\ref{figure:scale-ex} (a), a set of real values $X$ 
are mapped symmetrically and uniformly to
INTs $X_q \in [-2^{m-1}, 2^{m-1}-1]^k$ where $m$ is the
size of a two's complement number and $k$ is the number of values to be quantized together ($k=5$ in the figure).
A quantization scaling factor $s$ is applied to the $k$ values
such that they map into INTs after rounding ($X_q=RoundToInt(X/s$)).  
One method to compute a scaling factor is to align the maximum-observed value 
within the $k$ values to the largest representable INT: 
\begin{equation}
s = \frac{Max(X)}{2^{m-1}-1}.
\end{equation}
The overall quantization function is thus:
\begin{equation}
    X_q = Q(X, s) = RoundToInt(X/s).
\end{equation}
Performing the descaling step, $sX_q$, restores quantized values into the original scale.

\vspace{5pt}\noindent\textbf{Scaling Strategies.}
As illustrated in Figure~\ref{figure:scale-ex} (a)-(b), there are various methods for encoding
and computing with scale factors, including the use of more hardware-efficient 
quantized power-of-two or INT representations.  The optimal scaling approach 
depends on factors such as implementation complexities and overheads, and has a 
first-order impact on quantization error (see QSNR definition in Section~\ref{section:design_space}).
Producing scaling factors for static values, such as pre-trained weights, is
relatively straightforward. However, handling dynamic 
activations and gradients with data-dependent numerical distributions is more challenging. 
Thus it is necessary to employ strategies such as assigning conservative scaling factors to guard
against dynamic outliers or utilizing dynamic scaling approaches that update
the scaling factor as distributions change. 

\vspace{5pt}\noindent\textbf{Coarse-grained Scaling Support in SW.} Many software-based scaling strategies have been proposed in the literature, as surveyed in ~\cite{gholami2021survey}.
Partitioning an initial set of values into multiple sets and introducing multiple scaling factors, 
as shown in the example of Figure~\ref{figure:scale-ex} (c), is
an effective approach to mitigate outliers and reduce quantization error.
Unfortunately, today's software-based approaches are limited to coarse-grain scaling factors 
that must be amortized efficiently across a large number of values to be quantized 
(referred to as block granularity $k$ throughout this paper)~\cite{shen2020q}.  In practice, most software-based INT scaling methods in modern GPUs 
require $\sim 1K$ elements to balance efficiency and model accuracy. Larger values of $k$, 
coupled with the limitations of maintaining scaling factors at low overheads, also 
limit the opportunity to reduce quantization bit-width, which is needed to provide enough resolution
to encode a large set of quantized values while minimizing quantization error.

\vspace{5pt}\noindent\textbf{Fine-grained Scaling Support in HW.}
Block Floating-Point (BFP)-inspired approaches such as MSFP~\cite{darvish2020pushing} and Hybrid BFP~\cite{drumond2018hbfp}
diminish the impact of outliers by introducing dedicated hardware support for fine-grained
scaling on the order of $k \sim 10$ that reduces or eliminates the need for software-based heuristics.
BFP is closely related to INT-based quantization based on the same principles, with the exception
that the scaling factors are constrained to powers of two, which
are practical to implement in hardware at a fine granularity (typically within a single dot product unit).

\begin{figure}[ht]
   \centering
   \includegraphics[width=\columnwidth]{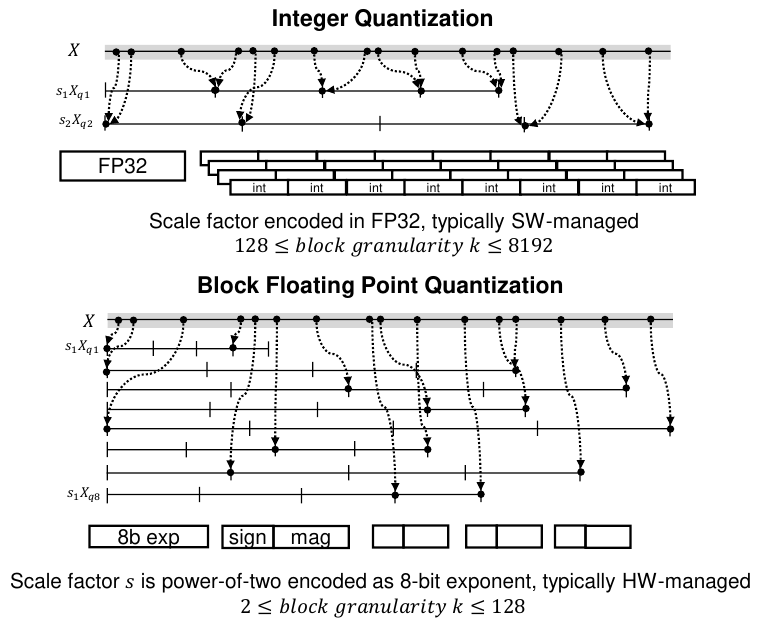}
    \caption{BFP and INT-based quantization operate on similar principles, but BFP offers superior resolution and numeric fidelity through fine-grained, hardware-managed scaling factors.}
    \label{figure:onelevelcomp}
    \vspace{-0.75em}
\end{figure}

Figure~\ref{figure:onelevelcomp} gives a visual intuition of quantization 
implemented with coarse-grained software versus fine-grained HW-based BFP. While both approaches 
operate on the same principles, the offloading of fine-grain scaling into HW 
enables much higher effective resolution in BFP vs. INT.

\section{Multi-level Scaling}
In this section, we extend the characterization of the aforementioned quantization and scaling approaches 
to newer techniques such as FP8, under the umbrella framework 
of BDR.
As implied in Section~\ref{sec:quantintro}, various hardware/software quantization approaches 
must balance block granularity ($k$) and the complexity and cost of handling scale factors.
Given a finite number of bits, the objective is to construct a high-fidelity representation
that minimizes computational and memory overheads and closely approximates full-precision scaling factors. 
In general, we expect quantization error
to decrease with smaller $k$ granularity and increased precision of scaling factors 
(e.g., FP32 vs. power-of-two). Decreasing $k$ and/or increasing the precision of scale factors 
adds to the cost of implementation and encoding.

\begin{figure}[ht]
   \centering
   \includegraphics[width=2.8in]{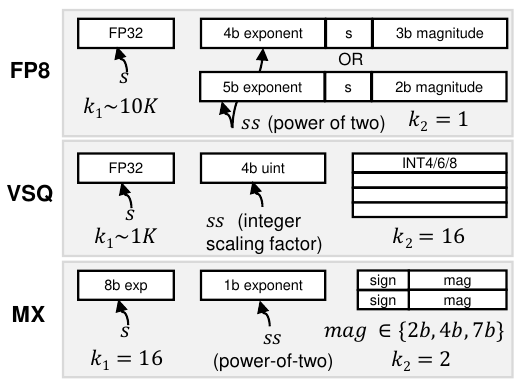}
    \caption{Three examples of two-level scaling with wide differences. 
    The MX format (this work) uses extremely fine-grain sub-scale factors in the form of hardware-managed 
    shared microexponents to reduce numeric noise at low silicon cost.}
    \label{figure:twolevelcomp}
    \vspace{-0.4em}
\end{figure}

As illustrated by the example in Figure~\ref{figure:twolevelex}, 
a two-level scaling approach enables an approximation of more accurate full-precision scaling.
A set of initial values are globally scaled using an ``expensive'' (amortized over $k$) FP32 rescaling step
based on the data distribution.  
Pre-scaled partitions $X_1$ and $X_2$ are then formed and can be further scaled ``cheaply'' using 
scaling factors $ss_1$ and $ss_2$ that are constrained to low-cost, power-of-two representation.
Comparisons of formats that leverage a two-level scaling approach, as shown in Figure~\ref{figure:twolevelcomp}, 
expose a larger design space than previously seen with INT- and BFP-based approaches. 
Table~\ref{tab:allcomp} summarizes all major techniques and how they differ in 
choices of encoding scaling ($s$) and sub-scaling ($ss$) factors, and block granularity ($k_1$, $k_2$).

\vspace{5pt}\noindent\textbf{Narrow-precision Floating-Point (FP8).}
The FP8 quantization approach,
recently popularized by Nvidia~\cite{nvidia2022fp8,transformerengine}, relies on a hybrid of hardware-software hierarchical scaling.
The FP8-based approach can be interpreted as a two-level scaling approach, where the high-level scale factor $s$,
is maintained by the Transformer Engine (a statistical-based heuristic for estimating scale factors) 
over a coarse granularity such as a tensor (up to $\sim 10K$), while 
compact sub-scale factors per element (effectively at $k_2=1$) are constrained
efficiently to powers of two that are privately stored for each element.

\vspace{5pt}\noindent\textbf{Per-Vector Scaled Quantization (VSQ).}
Another variant of two-level scaling is VSQ~\cite{dai2021vs}, which leverages per-vector integer scale factors within accelerator hardware. VSQ uses a coarse-grained software-driven scaling factor with $k_1 \sim 1K$, while also applying
a second-level INT-encoded scaling factor $ss_i$ over $16$ elements of INT4,
INT6, or INT8. This approach requires additional logic to handle
integer rescaling at a fine granularity within an AI accelerator's dot product unit.

\vspace{5pt}\noindent\textbf{Shared Microexponents.}
An extreme variant based on the concept of shared microexponents (MX), 
which Section~\ref{section:design_space} proposes and evaluates in detail, 
occupies a new point in the two-level design space with extremely granular 
parameters $k_1 = 16$ and $k_2 = 2$ with very narrow mantissa as low as two bits. 
As explained in Section~\ref{section:design_space}, variants of MX offer superior numerical 
fidelity normalized to silicon cost over previously proposed two-level schemes.

\begin{table}[]
\caption{Unification and classification of multiple quantization approaches under the two-level scaling framework. In our notation, $z\in \mathcal{Z}$ and $2^z$ indicates powers of two.}
\label{tab:allcomp}
\resizebox{\columnwidth}{!}{
\begin{tabular}{l|c|c|c|c|c|c|}
\cline{2-7}
\multicolumn{1}{l|}{} &
  {Scale} &
  {Sub-scale} &
  {$s$ type} &
  {$ss_i$ type} &
  {$k_1$} &
  {$k_2$} \\ \hline
\multicolumn{1}{|l|}{{INT}}      & SW & -  & FP32                   & -                      & $\sim 1K$ & -  \\ \hline
\multicolumn{1}{|l|}{{MSFP/BFP}} & HW & -  & $2^z$ & -                      & $\sim 10$ & -  \\ \hline
\multicolumn{1}{|l|}{{FP8}}      & SW & HW & FP32                   & $2^z$ & $\sim 10K$ & $1$  \\ \hline
\multicolumn{1}{|l|}{{VSQ}}      & SW & HW & FP32                   & INT                    & $\sim 1K$ & $\sim 10$ \\ \hline
\multicolumn{1}{|l|}{{MX}} &
  HW &
  HW &
  $2^z$ &
  $2^z$ &
  $\sim 10$ &
  $\sim 1$ \\ \hline
\end{tabular}}
\vspace{-0.75em}
\end{table}

\begin{figure}[t!]
    \centering
    \includegraphics[width=3.3in]{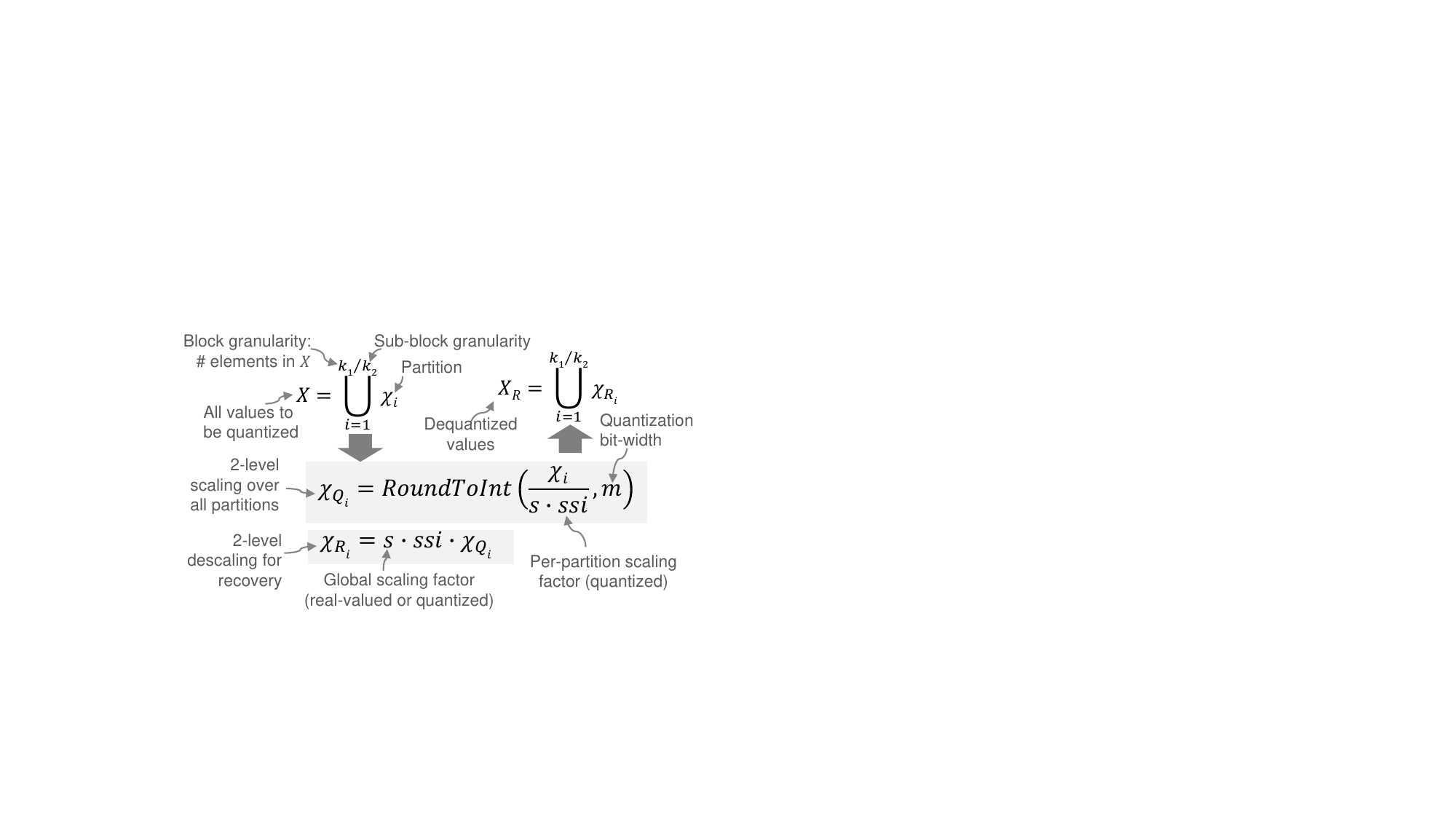}
    \caption{Two-level scaling and descaling with BDR.}
    \label{figure:bdrequation}
    \vspace{-0.75em}
\end{figure}

 \vspace{5pt}\noindent\textbf{Block Data Representations.} 
As illustrated in Figure~\ref{figure:bdrequation}, the BDR framework generalizes the two-level scaling examples by dividing a
global partition of size $k_1$ into multiple sub-partitions of equal-sized $k_2$
values. A global scaling factor $s$ is applied to all partitions for 
rescaling, while the sub-scale factors $ss_i$ are quantized for efficient
storage and computation and applied only to each respective local $i^{th}$ partition.

The total number of bits per element in BDR is $(m + 1) + d_1/k_1 + d_2/k_2$,
where $m$ represents the explicit mantissa bit-width\footnote{Scalar floating-point formats have an implicit leading $1$ by design. The variable $m$ in this formula does not account for this implicit bit.} and $d_1$ and $d_2$ represent the number of bits per individual scaling
factor $s$ and sub-scaling factor $ss_i$, respectively. 
Increasing $k_1$ and $k_2$ while decreasing $m$, $d_1$, and $d_2$
is desirable to reduce encoding and compute overheads.  However, the quantization noise
is minimized by increasing $m$, $d_1$, and $d_2$, while decreasing $k_1$ and $k_2$.
The dueling tensions across these sets of parameters are explored in Section~\ref{section:design_space}.

While outside the scope of this initial exploration, BDR can naturally extend beyond two levels, with the MX 
variants as prime candidates. Currently, MX selects compact powers-of-two 
representations for both the global ($s$) and local ($ss_i$) scaling factors that trades off scaling precision and flexibility for lowered implementation costs. 
This scheme could be improved further by introducing an even higher-level parent global scaling factor 
in software using high-precision FP32 scaling factors over an even coarser granularity at up to $\sim 1K$. 
MX also currently does not incorporate non-uniform approaches such as log-based representations or lookup
tables, or the use of other quantization formats for scaling factors. These explorations are left for future work. 

\section{Design Space Exploration}\label{section:design_space}
In this section, we first formalize the design space as a function of numerical fidelity and hardware cost. 
We then conduct an exhaustive sweep of design variables consisting of $800^+$ configurations to identify the Pareto-optimal data points.

\subsection{Numerical Fidelity}\label{subsec:numerical_fidelity}
To evaluate the numerical fidelity of BDR configurations, we adopt two approaches. The first is a statistical 
analysis to investigate the sensitivity of BDR to each design variable and how they contribute to the overall numerical 
robustness under different data distributions. This allows us to conduct an extensive search over the design space without 
running costly experiments for every configuration.

In addition to the statistical analysis, we conduct end-to-end training of various scale language models using different BDR 
configurations. We find a strong Pearson correlation \cite[Ch. 5.1]{tabachnick2007using} between the results 
of our statistical analysis and the language model loss achieved in our end-to-end training runs in the narrow bit-width 
regime. This suggests that our statistical results are a reliable predictor of the relative numerical fidelity performance 
of different configurations in practice at single-digit bit-widths.

In our statistical analysis, we use Quantization Signal to Noise Ratio~(QSNR) as a measure of numerical fidelity under different quantization schemes. QSNR is defined as the ratio of the power of the non-quantized signal (i.e., the original vector $\mathbf{X} = [x_1, x_2, \cdots, x_k] \in \mathbb{R}^{k}$) to the 
power of the quantization noise expressed in decibels. It is calculated as:
 \begin{equation}  \label{eq:qsnr_def}
 \resizebox{0.6\linewidth}{!}{$ 
\begin{aligned}
\text{QSNR} &:= -10 \, \text{log} (\frac{\mathbb{E}[\lVert Q(\mathbf{X}) - \mathbf{X} \rVert^{2}]}{\mathbb{E}[\lVert \mathbf{X} \rVert^{2}]}),
\end{aligned}$}
\end{equation}
where $\lVert \mathbf{X} \rVert := (\sum_{i=1}^{k}{x_{i}^{2}})^{1/2}$, and $Q(\cdot)$ is the quantization function which maps $k$-dimensional 
arrays with FP32 precision to the target quantization format. A higher QSNR indicates that the 
quantized vector better preserves the direction and magnitude of the original non-quantized vector in the space $\mathbb{R}^{k}$. The required QSNR level for a successful run varies depending on the task complexity, dataset, and model topology. We explore the trade-off between hardware efficiency and QSNR of different design points in Section~\ref{subsec:Pareto}.

\subsection{Hardware Cost}\label{subsec:hardware}
In this section, we outline a hardware architecture that supports operations on BDR formats and describes our methodology for estimating the costs of 
these operations in circuit area, power consumption, and memory.

\begin{figure}
	\includegraphics[width=0.9\columnwidth]{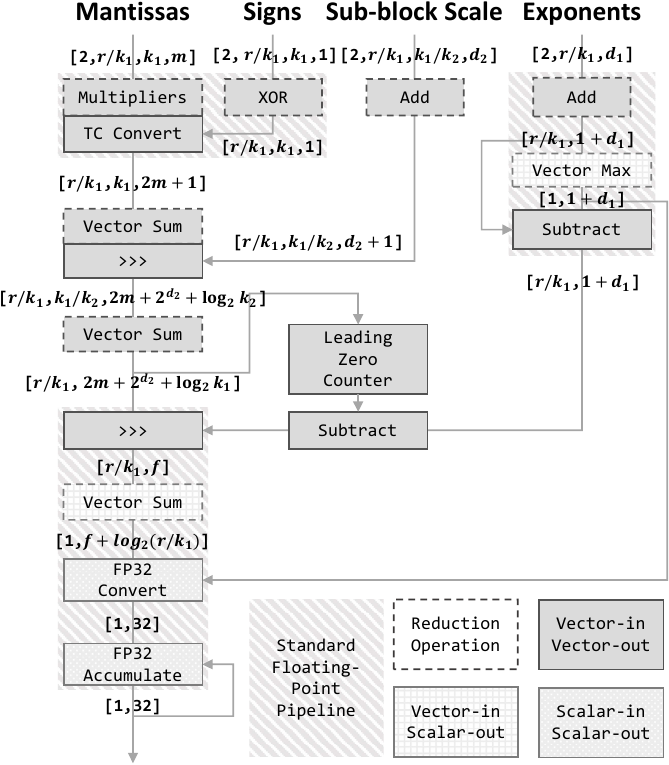}
	\caption{Typical hardware dot product pipeline used to model variants under the BDR model such as FP, BFP, and MX. We use a separate pipeline (not shown here due to space limitations) for settings that require a second-level INT-based scaling (e.g., VSQ). $r$ represents the reduction dimension of the dot product.
	Changing the parameters of this pipeline also implements conventional scalar floating-point~($k_1=k_2=1$)
	and block floating-point~($d_2=0$) dot product. The annotations in brackets represent the
	shape of each input or output to each hardware block. The right-most dimension is the number of bits.
	In our evaluations, we select $f$ to be the smaller of $25$ or the maximum possible dynamic range for each format.}
	\label{figure:dot_product_pipeline}
	\vspace{-1.8em}
\end{figure}

\vspace{5pt}\noindent\textbf{Dot Product Architecture}. Figure~\ref{figure:dot_product_pipeline} shows the hardware architecture of a floating-point dot product unit, which can be configured to support various scalar floating-point and MX formats. In addition to the parameters that define the 
BDR variants of interest (referred to as MX in Table~\ref{table:basic_formats}), the dot product unit in Figure~\ref{figure:dot_product_pipeline} is also parameterized by the dot product length $r$ and the fixed-point reduction accumulation precision $f$. When the block granularity is greater than $1$ (i.e., $k_1>1$), the first part of the pipeline reduces the Hadamard product of $k_1$ elements to a single element. There are $r/k_1$ such reductions. The results of these $r/k_1$ reductions are then normalized to the largest element and reduced in fixed-point in the second half of the pipeline. When the block is further broken down to a second-level of granularity (i.e., $k_2>1$), the pipeline performs a conditional right-shift, up to $2^{d_2} - 1$ bits, at the depth of $log_2(k_2)$ while summing the $k_1$ elements. 

In Figure~\ref{figure:dot_product_pipeline}, when the pipeline is configured with $k_1=k_2=1$, it implements a standard scalar floating-point dot product.
In this setting, the internal accumulation across the native reduction dimension is performed by normalizing the elements of a Hadamard product to the largest element and reducing in fixed-point. This is an optimistic approximation for scalar floating-point area overhead given that the data is serially accumulating in a floating-point type $f-2$ mantissa bits instead of building a full floating-point adder tree.

\begin{figure*}
\centering
	\includegraphics[width=0.9\textwidth]{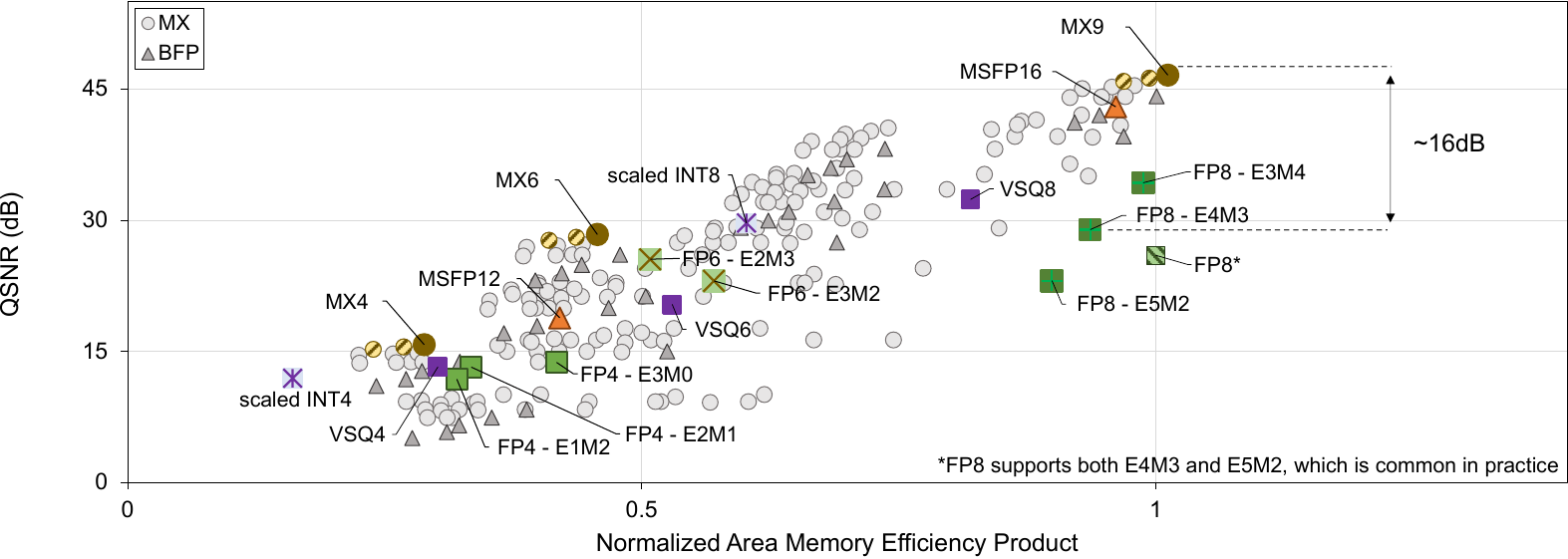}
	\caption{QSNR vs. the area-memory efficiency product for different BDR configurations. We focus our search space on symmetric dot product units.
	The area is normalized to a $64$-element FP8 dot product and the memory efficiency is the inverse of the packing efficiency of a $256$-element tile into a $64B$ memory interface. We chose a data distribution ($\mathbf{X} \sim \mathcal{N}(0,|\mathcal{N}(0,1)|)$) that is commonly found in DNN workloads. The conventional block floating-point data format is a subset of BDR with $d_2=0$ and is therefore a
	part of our BDR parameter search. MSFP12 and MSFP16 are instances of the conventional block floating-point proposed in~\cite{darvish2020pushing}.
	For integer (INT), VSQ~\cite{dai2021vs}, and scalar Floating-Point (FP) formats, we used FP32 first-level scaling factor
	based on the absolute maximum value over a window of past observed vectors similar to the ``delayed scaling'' approach proposed \cite{transformerengine} for scaling dynamic tensors with varied distribution during training. The use of offline scaling approaches (typical for inference) might shift the QSNR value of certain configurations. VSQ~\cite{dai2021vs} variants shown in this figure are the best of $d_2 = \{4, 6, 8, 10\}$.}
	\label{figure:qsnr_vs_area}
	\vspace{-15pt}
\end{figure*}

\vspace{5pt}\noindent\textbf{Synthesis and Area Estimation Methodology}. In the process of ASIC synthesis and physical design, various parameters can affect the circuit area and power consumption. For instance, synthesis tools can map adders and other blocks into circuits that reduce critical path delay at a cost of increased area~\cite{cmosvlsidesign}. In addition, multi-format dot product units in commercial implementations may share some sub-circuits across different data formats to save on circuit area. 

The BDR format is designed to be compatible with existing standard floating-point dot product pipelines in order to enable sub-circuit sharing and lower marginal costs. However, the cost of integrating BDR into a given design will depend on various factors, such as the underlying architecture, the set of data formats, and the relative throughputs of the existing design. To make the results of our synthesis and area estimation methodology more comparable, we normalize the standard-cell area of each BDR configuration to a baseline configurable FP8 dot product unit supporting both E4M3 and E5M2. This allows us to directly compare the relative costs of different BDR configurations and understand the trade-offs between numerical fidelity and hardware cost.

Due to software license usage restrictions, building hundreds of configurations and finding the optimal pipelining for each of them was not feasible. Our research on optimal register placements in smaller sweeps, however, shows that registers typically only accounted for about $10\%$ of the total area. Therefore, a design requiring half as many pipeline stages would only save around $5\%$ of the core area. To isolate the core area for a given configuration from differences in optimal pipeline stage placement or sub-circuit synthesis mapping, we synthesize each configuration with an easily achievable timing constraint of $10ns$ and with only inputs and outputs being registered. This ensures that synthesis implementation selection targets the minimum area in all designs. We use Synopsys Design Compiler and a leading-edge process node.

\vspace{5pt}\noindent \textbf{Memory Footprint}. 
While it is important to consider the area cost of the dot product unit, we should also consider the memory efficiency --- both from a capacity and bandwidth perspective. 
In particular, the DRAM or High Bandwidth Memory (HBM) interfaces are of a fixed width, therefore while accessing the tensors if the
data cannot be packed into the memory interface width, the effective memory capacity and/or bandwidth will be reduced which may result in lower performance (especially for inference use cases). Typical hardware architectures utilize tiling of the input tensors to achieve high performance on AI workloads. Therefore, for the memory footprint analysis, we consider the packing efficiency of a typical tile size of $256$~elements (although this is a hardware architecture design variable) into a $64B$ memory interface.

\subsection{Pareto-frontier formats}\label{subsec:Pareto}
Figure~\ref{figure:qsnr_vs_area} illustrates the trade-off between numerical fidelity and the hardware cost for different data format configurations. The x-axis indicates the implementation cost of each configuration as a function of the dot product area and memory efficiency. The reported QSNR (in decibels) on the y-axis is the averaged measured QSNR from over $10K$ independent vectors drawn from a normal Gaussian distribution with a variable variance (higher QSNR is better). We chose this distribution to cover a range of variances observed in gradient, error, weight, and activation tensors in a typical training cycle.

In our efficiency analysis, we normalize the dot product area to that of a baseline FP8 (supporting both E4M3 and E5M2) implementation and calculate the memory efficiency as the inverse of the ratio of the number of elements that can be packed into a $64B$ memory interface. Since the goal is to select a congruent data format for both training and inference, we give equal weight to the dot product area and memory efficiency as they are both important from a performance and power perspective for the two scenarios.

\vspace{5pt}\noindent \textbf{BDR QSNR Lower-Bound}. In Figure~\ref{figure:qsnr_vs_area}, we analyzed  the QSNR of various BDR configurations for a Gaussian distribution with varying variance.
Theorem 1 asserts a lower-bound on the QSNR of BDR formats as a function of 
the number of mantissa bits $m$, the block granularity $k_1$ and $k_2$, and scale bit-width $d_1$ and $d_2$.

\begin{theorem} \label{theorem:MSFPp_worst_case_lower_bound}
Given an N-dimensional vector $\mathbf{X}$ drawn from an arbitrary 
distribution $\rho$ in FP32 precision, the QSNR of the BDR-quantized version of $\mathbf{X}$ is lower-bounded by the following. Variable $d_1$ is set to $8$ for BDR data format and $\beta$ is an intermediate variable equal to $2^{d_2}-1$:\footnote{In general terms, $d_2$ denotes the minimum number of required bits to represent the difference between the global scaling factor $s$ and the smallest sub-scale factor among $ss_i$'s.}
	\begin{equation}  \label{eq:MSFPp_worst_case_lower_bound}
	\resizebox{0.9\linewidth}{!}{$\begin{aligned}
	\text{QSNR} & \geq 6.02 \, m + 10 \, \text{log} \, \big(\frac{2^{2\beta}}{\text{min}(N,k_1)+(2^{2\beta}-1)k_2}\big).
	\end{aligned}$}
	\end{equation}
\end{theorem}

A detailed proof of Theorem 1 can be found in Section~\ref{Sec:proof}. As also observed empirically in Figure~\ref{figure:qsnr_vs_area}, the QSNR has a linear relation with the number of mantissa bits $m$ and a logarithmic relation with the block granularity $k_1$ and $k_2$. Note that Eq.~\eqref{eq:MSFPp_worst_case_lower_bound} provides a parameterized lower-bound for the BDR family of data formats. This in turn provides a guarantee of the worst-case numerical fidelity of BDR formats in the face of skewed distributions with correlated noise. 

\vspace{5pt} \noindent \textbf{Basic Data Formats}.
To evaluate the efficacy of BDR data formats on AI workloads in the rest of this paper, we select three representative points from the Pareto frontier in Figure~\ref{figure:qsnr_vs_area}. The selected configurations are defined in Table~\ref{table:basic_formats} and are referred to as MX9, MX6, and MX4. We select a format (MX6) whose QSNR is comparable to that of FP8 but with approximately $2\times$ lower area-memory cost. We also select two formats whose QSNR are roughly $50\%$ higher (MX9) and $50\%$ lower (MX4) compared to that of FP8, with comparable and $4\times$ lower area-memory cost, respectively.

The specific parameters of these formats (see Table~\ref{table:basic_formats}) are chosen at the ``knee'' (shown as hashed points on the Pareto frontier in Figure~\ref{figure:qsnr_vs_area}) of a particular parameter's sweep by trading off the increase in QSNR with the increase in the normalized area-memory efficiency product. For instance, across all the selected formats, increasing the value of $d_2$ from $1$-bit to $2$-bit increases the QSNR by only~$0.5dB$, but with a~$30-50\%$ increase in the normalized cost. Similarly, with a $d_2$ value of $1$, reducing $k_2$ from $8$ to $2$ increases the QSNR by approximately~$2dB$, with only a marginal~$3\%$ increase in the normalized cost. Whereas, further reducing the value of $k_2$ from $2$ to $1$ increases the QSNR only by~$0.7dB$, but with a significant~$30-40\%$ increase in the normalized cost. We pragmatically select a first-level block granularity of $16$ to avoid forcing a large reduction dimension on hardware architectures. In general, we also observe that the trade-off between the increase in the QSNR versus the increase in the normalized area-memory efficiency product reduces as we increase the number of bits per element.

\begin{table}[ht]
\centering
  \caption{Definition of three basic MX data formats. Both scaling factors $d_1$ and $d_2$ are encoded as power-of-two exponents.}
  \label{table:basic_formats}
   \renewcommand{\arraystretch}{1} 
\begin{tabular}{llc|c|c|c|}
\cline{4-6}
\multicolumn{2}{l}{}                                                      &    & MX9 & MX6 & MX4 \\ \hline
\multicolumn{1}{|l|}{\multirow{2}{*}{Block granularity}} & \multicolumn{1}{l|}{1st level} & $k_1$ & 16 & 16 & 16 \\ \cline{2-6} 
\multicolumn{1}{|l|}{}                   & \multicolumn{1}{l|}{2nd level} & $k_2$ & 2    & 2    & 2    \\ \hline
\multicolumn{1}{|l|}{\multirow{2}{*}{Scale bit-width}}   & \multicolumn{1}{l|}{1st level} & $d_1$ & 8  & 8  & 8  \\ \cline{2-6} 
\multicolumn{1}{|l|}{}                   & \multicolumn{1}{l|}{2nd level} & $d_2$ & 1    & 1    & 1    \\ \hline
\multicolumn{1}{|l|}{Mantissa bit-width} & \multicolumn{2}{c|}{$m$}              & 7    & 4    & 2    \\ \hline
\multicolumn{3}{|l|}{Average bits per element}                                    & 9    & 6    & 4    \\ \hline
\end{tabular}
\renewcommand{\arraystretch}{1}
\vspace{-3pt}
\end{table}

Even though MX9 is not necessarily on the Pareto frontier in Figure~\ref{figure:qsnr_vs_area}, it is still a good choice for the basic data format for further evaluation as it allows all the selected formats to have the same settings (see Table~\ref{table:basic_formats}) except for their mantissa bit-widths, which allows for maximum hardware reuse. Moreover, the difference in QSNR~($\approx0.8dB$) between the selected MX9 format and the configuration on the Pareto frontier is insignificant for the efficacy of the end-to-end training and inference workloads due to its already high QSNR value. Note that MX9 has approximately $3.6dB$ higher QSNR compared to that of MSFP16~\cite{darvish2020pushing}. Empirical measurements of proprietary generative inference models (results not reported) show a significant accuracy degradation in MSFP16 as a drop-in replacement option that is not occurring in MX9.

MX9 has a hardware efficiency close to that of FP8 with significantly higher numerical fidelity. For instance, in the case of a Gaussian distribution with variable variance, the QSNR of MX9 is about $16dB$ higher than FP8~(E4M3). A $16dB$ higher fidelity is roughly equivalent to having $2$ more mantissa bits in the scalar floating-point format. For the same distribution, MX6's QSNR lies between the two FP8 variants E4M3 and E5M2 while providing an approximately $2\times$ advantage on the hardware cost as measured by the normalized area memory efficiency product. We qualitatively observed a similar trend under different data distributions as well.

\section{Compute Flow} \label{sec:compute_flow}
In this section, we describe the computation process for training and inferencing with the MX data format.

\begin{table*}[]
  \begin{minipage}{\textwidth}
    \vspace{8pt}
    \begin{center}
    \caption{Training and inferencing with MX data formats. MX9 can be used as a drop-in replacement for high-precision data formats like FP32 or BF16 in the training and inferencing pipeline without the dependency on complex online statistical heuristics or any change in the hyper-parameter settings. The results of direct casting to MX6 format are also included in this table for completeness. The delta in MX training accuracies (both higher and lower) are typically within the run-to-run variation of FP32-based training when initialized with different random seeds or trained in different containers. The baseline accuracy numbers reported in this table are the FP32 accuracy obtained under the same setting in which MX training is run (exact same seed, container, and node) and not the averaged accuracy under multiple runs. MX6 is expected to provide roughly $2\times$ improvement over FP8 in terms of area-memory efficiency and achieves inferences results close to FP32 accuracy with a modest level of quantization-aware fine-tuning.}
    \label{table:benchmark_results_v2}
    \renewcommand{\arraystretch}{1.1}
  \resizebox{\textwidth}{!}{\begin{tabular}{lllllccccc|}
  \cline{6-10}
     &
     &
     &
     &
    \multicolumn{1}{l|}{} &
    \multicolumn{2}{c|}{Training} &
    \multicolumn{3}{c|}{Inferencing} \\ \hline
    \multicolumn{1}{|l|}{Task} &
    \multicolumn{1}{l|}{Family} &
    \multicolumn{1}{l|}{Model} &
    \multicolumn{1}{l|}{Dataset} &
    \multicolumn{1}{l|}{Metric} &
    \multicolumn{1}{c|}{Baseline FP32} &
    \multicolumn{1}{c|}{MX9} &
    \multicolumn{1}{c|}{\begin{tabular}[c]{@{}c@{}}Direct Cast \\ (MX9)\end{tabular}} &
    \multicolumn{1}{c|}{\begin{tabular}[c]{@{}c@{}}Direct Cast\\ (MX6)\end{tabular}} &
    \begin{tabular}[c]{@{}c@{}}QA Fine-tuning\\ (MX6)\end{tabular} \\ \hline
    \multicolumn{1}{|l|}{\multirow{3}{*}{\begin{tabular}[c]{@{}l@{}} Language \\ Translation \end{tabular}} } &
    \multicolumn{1}{l|}{\multirow{2}{*}{\begin{tabular}[c]{@{}l@{}}Transformer\\ (Enc-Dec)\end{tabular}}} &
    \multicolumn{1}{l|}{Transformer-Base~\cite{transformer}} &
    \multicolumn{1}{l|}{\multirow{2}{*}{WMT-17~\cite{wmt17}}} &
    \multicolumn{1}{l|}{\multirow{2}{*}{BLEU Score~$\uparrow$}} &
    \multicolumn{1}{c|}{26.85} &
    \multicolumn{1}{c|}{26.51} &
    \multicolumn{1}{c|}{26.55} &
    \multicolumn{1}{c|}{26.32} &
    26.81 \\ \cline{3-3} \cline{6-10} 
    \multicolumn{1}{|l|}{} &
    \multicolumn{1}{l|}{} &
    \multicolumn{1}{l|}{Transformer-Large~\cite{transformer}} &
    \multicolumn{1}{l|}{} &
    \multicolumn{1}{l|}{} &
    \multicolumn{1}{c|}{27.63} &
    \multicolumn{1}{c|}{27.77} &
    \multicolumn{1}{c|}{27.60} &
    \multicolumn{1}{c|}{27.48} &
    27.62 \\ \cline{2-4} \cline{6-10}
    \multicolumn{1}{|l|}{} &
    \multicolumn{1}{l|}{LSTM} &
    \multicolumn{1}{l|}{GNMT~\cite{gnmt}} &
    \multicolumn{1}{l|}{WMT-16~\cite{wmt16}} &
    \multicolumn{1}{l|}{} &
    \multicolumn{1}{c|}{24.44} &
    \multicolumn{1}{c|}{24.47} &
    \multicolumn{1}{c|}{24.45} &
    \multicolumn{1}{c|}{24.45} &
    - \\ \cline{1-10}
    \multicolumn{1}{|l|}{\begin{tabular}[c]{@{}l@{}} Language \\ Encoding \end{tabular}} &
    \multicolumn{1}{l|}{\multirow{2}{*}{\begin{tabular}[c]{@{}l@{}}Transformer\\ (Enc-Only)\end{tabular}}} &
    \multicolumn{1}{l|}{BERT-Base~\cite{gptbert}} &
    \multicolumn{1}{l|}{\multirow{2}{*}{Wikipedia~\cite{wiki}}} &
    \multicolumn{1}{l|}{\multirow{2}{*}{\begin{tabular}[c]{@{}l@{}}Perplexity $\downarrow$ \end{tabular}}} &
    \multicolumn{1}{c|}{4.58} &
    \multicolumn{1}{c|}{4.62} &
    \multicolumn{3}{c|}{\multirow{2}{*}{See Table~\ref{table:bert_acc} for detail.}} \\ \cline{3-3} \cline{6-7} 
    \multicolumn{1}{|l|}{} &
    \multicolumn{1}{l|}{} &
    \multicolumn{1}{l|}{BERT-Large~\cite{gptbert}} &
    \multicolumn{1}{l|}{} &
    \multicolumn{1}{l|}{} &
    \multicolumn{1}{c|}{3.58} &
    \multicolumn{1}{c|}{3.58} &
    \multicolumn{1}{c}{} &
    \multicolumn{1}{c}{} &
     \\ \cline{1-10} 
    \multicolumn{1}{|l|}{\begin{tabular}[c]{@{}l@{}} Language \\ Modeling \end{tabular}}  &
    \multicolumn{1}{l|}{\begin{tabular}[c]{@{}l@{}}GPT\\ (Dec-Only)\end{tabular}}&
    \multicolumn{8}{c|}{See tables~\ref{table:generative_training} and~\ref{table:generative_inferencing} for detail.} \\ \cline{1-10} 
    \multicolumn{1}{|l|}{\multirow{5}{*}{\begin{tabular}[c]{@{}l@{}} Image \\ Classification \end{tabular}}} &
    \multicolumn{1}{l|}{\multirow{2}{*}{\begin{tabular}[c]{@{}l@{}} Vision \\ Transformer \end{tabular}} } &
    \multicolumn{1}{l|}{ DeiT-Tiny~\cite{deit}} &
    \multicolumn{1}{l|}{\multirow{5}{*}{\begin{tabular}[c]{@{}l@{}}ImageNet\\ ILSVRC12~\cite{imagenet12}\end{tabular}}} &
    \multicolumn{1}{l|}{\multirow{7}{*}{Top-1 Acc.~$\uparrow$}} &
    \multicolumn{1}{c|}{72.16\%} &
    \multicolumn{1}{c|}{72.84\%} &
    \multicolumn{1}{c|}{72.2\%} &
    \multicolumn{1}{c|}{71.23\%} &
    71.96\% \\ \cline{3-3} \cline{6-10} 
    \multicolumn{1}{|l|}{} &
    \multicolumn{1}{l|}{} &
    \multicolumn{1}{l|}{DeiT-Small~\cite{deit}} &
    \multicolumn{1}{l|}{} &
    \multicolumn{1}{l|}{} &
    \multicolumn{1}{c|}{80.53\%} &
    \multicolumn{1}{c|}{80.31\%} &
    \multicolumn{1}{c|}{80.52\%} &
    \multicolumn{1}{c|}{80.07\%} &
    80.34\% \\ \cline{2-3} \cline{6-10} 
    \multicolumn{1}{|l|}{} &
    \multicolumn{1}{l|}{\multirow{3}{*}{CNN}} &
    \multicolumn{1}{l|}{ResNet-18~\cite{resnet}} &
    \multicolumn{1}{l|}{} &
    \multicolumn{1}{l|}{} &
    \multicolumn{1}{c|}{70.79\%} &
    \multicolumn{1}{c|}{70.44\%} &
    \multicolumn{1}{c|}{70.80\%} &
    \multicolumn{1}{c|}{69.35\%} &
    70.74\% \\ \cline{3-3} \cline{6-10} 
    \multicolumn{1}{|l|}{} &
    \multicolumn{1}{l|}{} &
    \multicolumn{1}{l|}{ResNet-50 v1.5~\cite{resnet}} &
    \multicolumn{1}{l|}{} &
    \multicolumn{1}{l|}{} &
    \multicolumn{1}{c|}{77.41\%} &
    \multicolumn{1}{c|}{77.09\%} &
    \multicolumn{1}{c|}{77.16\%} &
    \multicolumn{1}{c|}{75.63\%} &
    77.00\% \\ \cline{3-3} \cline{6-10} 
    \multicolumn{1}{|l|}{} &
    \multicolumn{1}{l|}{} &
    \multicolumn{1}{l|}{MobileNet v2~\cite{mobilenet}} &
    \multicolumn{1}{l|}{} &
    \multicolumn{1}{l|}{} &
    \multicolumn{1}{c|}{72.14\%} &
    \multicolumn{1}{c|}{71.56\%} &
    \multicolumn{1}{c|}{71.48\%} &
    \multicolumn{1}{c|}{67.64\%} &
    71.25\% \\ \cline{1-4} \cline{5-10}
    \multicolumn{1}{|l|}{\multirow{2}{*}{\begin{tabular}[c]{@{}l@{}} Denoising \\ Diffusion \end{tabular}}} &
    \multicolumn{1}{l|}{\multirow{2}{*}{UNET}} &
    \multicolumn{1}{l|}{Conditioned DDPM~\cite{diffusion}} &
    \multicolumn{1}{l|}{\multirow{2}{*}{ImageNet-64~\cite{imagenet64}}} &
    \multicolumn{1}{l|}{\multirow{2}{*}{\begin{tabular}[c]{@{}l@{}} FID. $\downarrow$ $\mid$ \\ Inception Score $\uparrow$ \end{tabular}}} &
    \multicolumn{1}{c|}{7.60 $\mid$ 34.76} &
    \multicolumn{1}{c|}{5.37 $\mid$ 34.14} &
    \multicolumn{1}{c|}{7.81 $\mid$ 37.40} &
    \multicolumn{1}{c|}{26.62 $\mid$ 27.88} &
    15.72 $\mid$ 31.77 \footnote{\label{note1}FID score is an statistical metric used to compare the distribution of generated images compared to a reference set. FID is known to have a high variance sensitive to data, image size, etc. Inception score, on the other hand, evaluates the distribution of the generated images and has less variance.}
     \\ \cline{3-3} \cline{6-10} 
    \multicolumn{1}{|l|}{} &
    \multicolumn{1}{l|}{} &
    \multicolumn{1}{l|}{Unconditioned DDPM~\cite{diffusion}} &
     \multicolumn{1}{l|}{} &
    \multicolumn{1}{l|}{} &
    \multicolumn{1}{c|}{21.99 $\mid$ 15.34} &
    \multicolumn{1}{c|}{21.46 $\mid$ 15.72} &
    \multicolumn{1}{c|}{17.79 $\mid$ 15.83} &
    \multicolumn{1}{c|}{44.74 $\mid$ 13.10} &
    29.55 $\mid$ 15.47 
    \\ \hline
    \multicolumn{1}{|l|}{\multirow{1}{*}{Speech Rec.}} &
    \multicolumn{1}{l|}{\multirow{1}{*}{Transformer}} &
    \multicolumn{1}{l|}{Wav2Vec 2.0~\cite{wav2vec2}} &
    \multicolumn{1}{l|}{\multirow{1}{*}{Librispeech~\cite{librispeech}}} &
    \multicolumn{1}{l|}{\multirow{1}{*}{WER~$\downarrow$}} &
    \multicolumn{1}{c|}{18.9} &
    \multicolumn{1}{c|}{17.27} &
    \multicolumn{1}{c|}{18.94} &
    \multicolumn{1}{c|}{20.98} &
    20.13
  
     \\ \hline
    \multicolumn{1}{|l|}{Recommendation} &
    \multicolumn{1}{l|}{MLPs} &
    \multicolumn{1}{l|}{DLRM~\cite{dlrm}} &
    \multicolumn{1}{l|}{Criteo Terabyte~\cite{criteo}} &
    \multicolumn{1}{l|}{AUC~$\uparrow$} &
    \multicolumn{1}{c|}{0.8028} &
    \multicolumn{1}{c|}{0.8026} &
    \multicolumn{1}{c|}{0.8027} &
    \multicolumn{1}{c|}{0.8013} &
    - \\\hline
    \renewcommand{\arraystretch}{1}
	\vspace{-23pt}
  \end{tabular}}
\end{center}
\end{minipage}
	\vspace{-12pt}
  \end{table*}

\begin{figure}[ht]
\centering
    \includegraphics[width=\columnwidth]{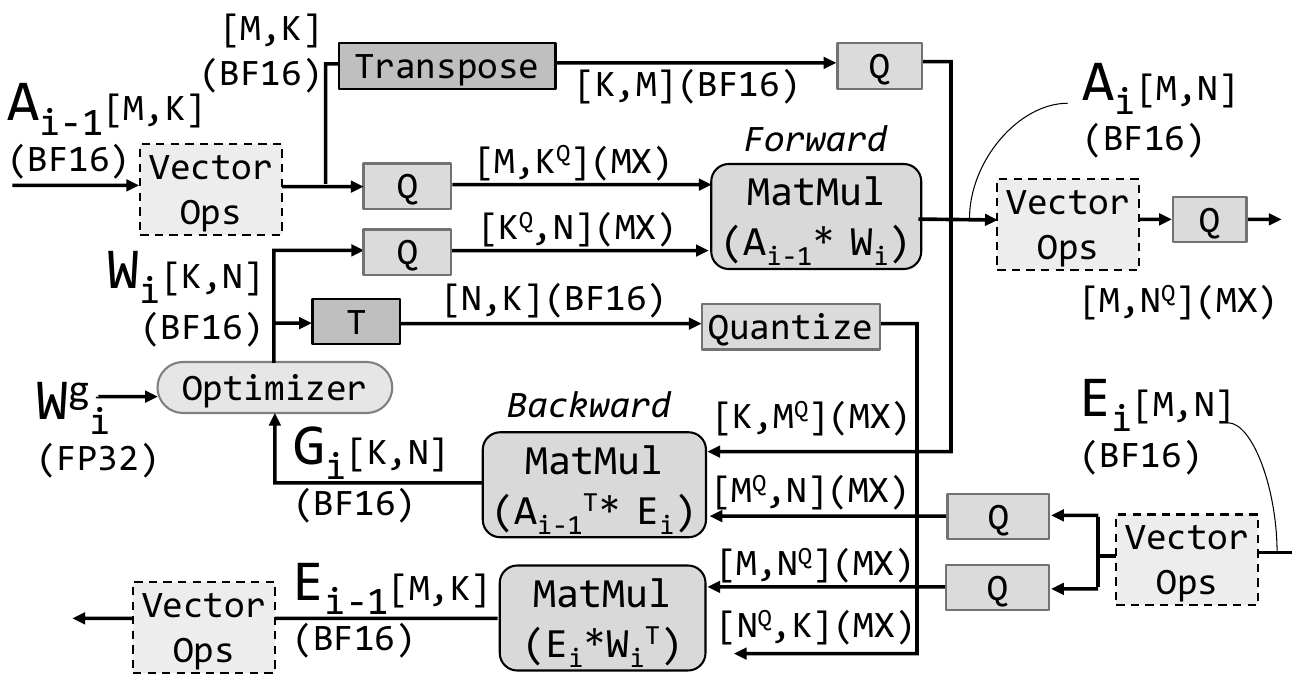}
    \caption{Compute flow graph of a training iteration including the forward and backward passes with MX quantization. The superscript \textquote{Q} in the dimensions 
    (e.g., ``MX[$K$,$M^Q$]'') indicates the axis along which the tensor is quantized. 
    }
    \label{figure:compute_flow}
\end{figure}

\noindent \textbf{Training}. Figure~\ref{figure:compute_flow} illustrates the computation flow of a training 
iteration using MX format. Tensor reduction operations, such as matrix multiplications and convolutions, are 
performed in MX during both the forward and backward passes. Both inputs of a tensor operation are MX-quantized. 
Element-wise operations, such as Layernorm, Softmax, GELU, and residual addition are performed in a scalar 
floating-point format like BF16~\cite{kalamkar2019study} or FP32. In our experiments, we use BF16 as the default data format for 
element-wise operations with the exception of the vector operations in the diffusion loop and the Softmax in the 
mixture-of-experts gating function~\cite{fedus2021switch}.

MX is a directional data format that requires tensors to be quantized along the reduction dimension to attain 
hardware benefits. This association with a particular dimension means that MX quantization and the transpose 
operation are not commutative. In the weight update stage, as shown in Figure~\ref{figure:compute_flow}, two 
quantized versions of the weights are generated: one for the forward pass and one for the backward pass. The 
transpose operation must be performed before quantizing the weights for the backward pass. Note that both copies do 
not need to be stored in working memory at the same time unless a very fine-grained interleaving of the forward and 
backward passes (or recompute strategy) is deployed. 

\vspace{1pt}
\noindent \textbf{Inferencing}. Narrow bit-width deployment is made easier with MX data format as training and 
inferencing use the same congruent data format. In Section~\ref{sec:results}, we also evaluate the use of MX for 
inferencing models trained in other formats. 
For post-training quantization, we 
consider two scenarios:

\vspace{1pt}
\noindent \textit{Direct cast (drop-in replacement)}. In this setting, we take a pre-trained model in higher 
precision (e.g., FP32), perform a straight cast into MX data format, and evaluate the 
model quality. 

\vspace{1pt}
\noindent \textit{Quantization-aware fine-tuning}. In this setting, we take a pre-trained model in higher precision 
(e.g., FP32), cast it into MX format, and perform a few iterations of quantization-aware 
fine-tuning to recover any potential accuracy drop. This approach is mostly used for inferencing with narrower data 
formats like MX6 and MX4. The compute flow for the fine-tuning step is similar to that shown in 
Figure~\ref{figure:compute_flow}, except that the forward and backward passes may use different bit-widths. For 
example, the forward pass might use MX6 or MX4 and the backward pass might use a higher bit-width format 
(e.g., MX9, or FP32).

In memory-intensive scenarios such as recommendation models, we can further improve inferencing performance by 
quantizing both storage and compute. In evaluating the DLRM model, we leverage this optimization by quantizing both the embedding tables and tensor 
computations to MX format.

\section{Empirical Results}\label{sec:results}

\noindent \textbf{Experimental Setup}. In our evaluations, we use a custom CUDA library to emulate MX data formats on current GPU architectures. Our custom library reproduces numerical results identical to what a native-MX silicon would produce. 
All tensor operations in forward and backward passes are performed with the MX data format (including the first and last 
layers) unless explicitly called out otherwise. 

\begin{table*}[tb]
  \begin{minipage}{\textwidth}
  \vspace{8pt}
\caption{Zero-shot and few-shot inferencing with MX data format on the OpenAI GPT3-175B model~\cite{browngpt3}. All the reported accuracies are the results of a ``\textbf{direct cast}'' into MX format with no quantization-aware fine-tuning. The tuple (w, a) indicates the format used for weights and activation. Note that few-shot accuracy in ANLI-r2 is lower than zero-shot accuracy due to the nature of this task.}
  \label{table:generative_inferencing}
  \renewcommand{\arraystretch}{1}
\resizebox{\textwidth}{!}{\begin{tabular}{|l|c|c|c|c|c|c|c|c|}
\hline
Task & N-shot & Baseline FP32 & (MX9, MX9) & (MX6, MX9) & (MX6, MX6) & (MX4, MX9) & (MX4, MX6) & (MX4, MX4) \\ \hline
\multirow{3}{*}{Hellaswag}  & 0 & $76.2\pm 0.4$ & $76.2\pm 0.4$ & $76.0\pm 0.4$ & $75.8\pm 0.4$ & $74.5\pm 0.4$ & $74.0\pm 0.4$ & $61.3\pm 0.5$ \\ \cline{2-9} 
                            & 1 & $76.5\pm 0.4$ & $76.6\pm 0.4$ & $76.3\pm 0.4$ & $76.0\pm 0.4$ & $74.0\pm 0.4$ & $73.6\pm 0.4$ & $59.1\pm 0.5$ \\ \cline{2-9} 
                            & 2 & $77.4\pm 0.4$ & $77.3\pm 0.4$ & $77.3\pm 0.4$ &  $76.8\pm 0.4$& $74.8\pm 0.4$ & $74.3\pm 0.4$ & $59.4\pm 0.4$  \\ \hline
\multirow{3}{*}{WIC}        & 0 & $48.3\pm 2.0$ & $47.8\pm 2.0$ & $48.0\pm 2.0$ & $48.1\pm 2.0$ & $48.9\pm 2.0$ & $48.0\pm 2.0$ & $50.3\pm 2.0$ \\ \cline{2-9} 
                            & 1 & $50.9\pm 2.0$ & $50.8\pm 2.0$ & $51.4\pm 2.0$ & $50.3\pm 2.0$ & $49.1\pm 2.0$ & $47.5\pm 2.0$ & $46.9 \pm 2.0$  \\ \cline{2-9} 
                            & 2 & $50.5\pm 2.0$ & $50.5\pm 2.0$ & $50.5\pm 2.0$ & $50.6\pm 2.0$ & $48.9\pm 2.0$ & $49.7\pm 2.0$ & $52.2\pm 2.0$  \\ \hline
\multirow{3}{*}{Anli-r2}    & 0 & $35.6\pm 1.6$ & $35.2\pm 1.5$ & $36.6\pm 1.5$ & $36.2\pm 1.5$ & $34.0\pm 1.5$ & $34.3\pm 1.5$ & $35.8\pm 1.5$ \\ \cline{2-9} 
                            & 1 & $33.8\pm 1.5$ & $33.7\pm 1.5$ & $33.4\pm 1.5$ & $32.8\pm 1.5$ & $33.0\pm 1.5$ & $33.3\pm 1.5$ & $31.7\pm 1.5$ \\ \cline{2-9} 
                            & 2 & $34.5\pm 1.5$ & $33.9\pm 1.5$ & $34.0\pm 1.5$ & $35.5\pm 1.5$ & $34.1\pm 1.5$ & $34.2\pm 1.5$ & $33.0\pm 1.5$ \\ \hline
\multirow{3}{*}{Winogrande} & 0 & $70.8\pm 1.3$ & $71.0\pm 1.3$ & $71.1\pm 1.3$ & $70.7\pm 1.3$ & $71.5\pm 1.3$ & $70.2\pm 1.3$ & $66.1\pm 1.3$ \\ \cline{2-9} 
                            & 1 & $73.0\pm 1.3$ & $72.2\pm 1.2$ & $71.6\pm 1.2$ & $71.5\pm 1.2$ & $71.4\pm 1.2$ & $71.4\pm 1.2$ & $64.1 \pm 1.2$ \\ \cline{2-9} 
                            & 2 & $74.0\pm 1.2$ & $74.3\pm 1.2$ & $73.8\pm 1.2$ & $74.9\pm 1.2$ & $72.9\pm 1.2$ & $72.5\pm 1.2$ & $61.7\pm 1.2$  \\ \hline
\end{tabular}}
\renewcommand{\arraystretch}{1}
\\
\end{minipage}
	\vspace{-20pt}
\end{table*}

\subsection{Benchmarks}\label{subsec:benchmark}
To evaluate the quality of MX data formats, we created a comprehensive benchmark suite featuring over $20$ 
representative open-source and proprietary AI models. In the following, we provide an overview of the models in our benchmark suite.

\vspace{5pt}
\noindent\textbf{Text and Natural Language}. Natural language processing applications can be roughly broken down into three subsets:

\begin{itemize}[leftmargin=*]
\vspace{3pt}
\item \noindent\textit{Neural Machine Translation}. We evaluate two model topologies in this category: LSTM-based and transformer-based. 
We use the MLPperf GNMT-V2 model~\cite{wu2016gmnt} as the LSTM benchmark. 
The transformer models are based on the work of~\cite{vaswani2017attention}. 
These models are trained using the WMT17-EN-DE data~\cite{bojar2017wmt} and 
evaluated on the WMT14-EN-DE~\cite{bojar2014wmt}.

\vspace{3pt}
\item \noindent\textit{Text Encoding and Language Understanding}. For this application, we evaluate the popular
BERT~\cite{devlin2018bert} models: BERT-base and BERT-large. 
The models are trained on the Wikipedia data and evaluated on SQuAD-v1.1~\cite{rajpurkar2016squad}. 

\vspace{3pt}
\item \noindent\textit{Generative Language Modeling}. Auto-regressive generative learning is probably the most significant and 
fastest-growing application in the NLP domain. In our benchmark suite, we evaluate MX performance on a variety of dense Generative Pre-Trained Transformers (GPT) models of size $6M$ up to $175B$. All models 
are trained with a sequence length of $1024$ to efficiency with the number of tokens processed estimated by GPT 
scaling power laws~\cite{kaplan2020scaling}. In addition to dense GPT models, we also evaluate  
an example Mixture-of-Experts (MoE) generative model from~\cite{rajbhandari2022deepspeed}. The MoE model in our 
study has $1.9B$ parameters trained for $40B$ tokens.
\end{itemize}

\vspace{3pt}
\noindent\textbf{Computer Vision}.  
Image classification and image generation are two important applications in computer vision:

\begin{itemize}[leftmargin=*]
\vspace{3pt}
\item \noindent\textit{Image Classification}. We evaluate three main families of models for image classification: ResNet~\cite{he2015resnet}, MobileNet-v2~\cite{sandler2018mobilenetv2}, and DeiT~\cite{touvron2021deit}. 
These benchmarks are trained and validated with ImageNet ILSVRC12 data~\cite{deng2009imagenet}.

\vspace{3pt}
\item \noindent\textit{Image Generation}. 
We use the DDPM model~\cite{nicholdiffusion} as a representative model in this category.  
We evaluate two variations of the DDPM model in our study. Both 
models are trained with $4000$ diffusion steps using ImageNet-64.
\end{itemize}

\vspace{3pt}
\noindent\textbf{Audio and Speech}. In this category, we evaluate the base configuration of Wave2Vec 2.0~\cite{baevski2020wav2vec2} using Librispeech data~\cite{panayotov2015libri}.

\vspace{3pt}
\noindent\textbf{Categorical Data}. One of the industrial applications of deep learning is processing categorical data in recommendation systems. In our benchmark suite, we include the MLperf Deep Learning Recommendation Model (DLRM)~\cite{naumov2019dlrm} in addition to three internal production models. We refer to these internal models as PR-rec1, PR-rec2, and PR-rec3.
The models are trained on large-scale internal data with tens or hundreds of billion samples collected from 
online recommendation services. PR-rec1 utilizes the canonical DLRM architecture at a much larger scale than open-source models, with nearly $1000$ embedding tables and over $100$ fully-connected layers. This model, with $32$-bit parameters, amounts to several terabytes in size. PR-rec2 and PR-rec3 use different interaction architectures 
compared to the DLRM, with PR-rec2 using transformer-based encoders and PR-rec3 using DHEN~\cite{Zhangdhen}.\footnote{The evaluation on the production recommendation models is done as part of a collaboration between Meta and Microsoft. All the evaluations on the rest of benchmarks are performed by Microsoft.}

\begin{table}[]
\caption{Question-Answering using BERT. Each entry reports {Exact Match / F1 score}. In both benchmarks, no quantization-aware fine-tuning is needed even with the MX6 data format.}
  \label{table:bert_acc}
  \renewcommand{\arraystretch}{1}
  \centering
{\begin{tabular}{|l|c|c|c|}
\hline
Model                & \begin{tabular}[c]{@{}c@{}}Baseline\\ FP32 \end{tabular} & \begin{tabular}[c]{@{}c@{}}Direct Cast\\ (MX9)\end{tabular} & \begin{tabular}[c]{@{}c@{}}Direct Cast\\ (MX6)\end{tabular}   \\ \hline
Bert-Base            & $80.80$ / $88.46$ & $80.71$ / $88.45$ & $80.62$ / $88.36$  \\ \hline
Bert-Large           & $87.65$ / $93.48$ & $87.63$ / $93.45$ & $87.49$ / $93.37$   \\ \hline
\end{tabular}}
\renewcommand{\arraystretch}{1}
\vspace{-8pt}
\end{table}

\subsection{Inferencing with the MX data format}\label{subsec:inference}
Tables~\ref{table:benchmark_results_v2}~to~\ref{table:bert_acc} summarize the MX inferencing results in both 
direct casting and quantization-aware fine-tuning settings across a variety of model topologies, scales, and 
modalities. As shown, a direct cast to MX9 can be mostly used as a drop-in replacement to high-precision formats. 

Direct casting to the more efficient MX6 format sometimes causes model accuracy to degrade. In these cases, the 
loss in accuracy can be recovered through a modest amount of quantization-aware fine-tuning. We report both the 
direct cast and fine-tuning results for MX6 in Tables~\ref{table:benchmark_results_v2}~and~\ref{table:bert_acc}. 
We generally saw the best quantization-aware fine-tuning results when we reset the optimizer, adjusted the initial 
learning rate, and eliminated rate decay, dropout, and momentum. 
In all our fine-tuning experiments, we use FP32 for the backward pass. The 
amount of fine-tuning necessary varied across benchmarks but was always much shorter than the original training duration.

The MX data format also presents opportunities for optimizing zero-shot and few-shot generative inferencing with 
no task-specific fine-tuning. See Table~\ref{table:generative_inferencing} for quantization results of OpenAI GPT3-175B model with MX formats. Consistent with the results for discriminative inferencing reported in 
Table~\ref{table:benchmark_results_v2}, MX9 provides a drop-in replacement alternative for efficient generative 
inferencing. Generative inferencing at large-scale allows the use of even narrower data formats such as MX4 and 
MX6 without the need for much quantization-aware fine-tuning. To the best of our knowledge, this is the first work 
reporting high-quality generative inferencing with both weights and activations quantized to ultra-narrow bit-width 
through a direct cast. In prior quantized generative inferencing work, the activation is typically kept at high 
precision BF16/FP16 and only weights are being quantized~\cite{dettmers2022case, pope2022efficiently}. 
This is because INT8/INT4 quantization of activations usually requires careful offline outlier suppression~\cite{xiao2022smoothquant} to achieve reasonable results.

\subsection{Training with MX data format}\label{subsec:training}
MX9 provides a robust drop-in replacement for high bit-width AI training (i.e., FP32/BF16/FP16) without 
requiring an outlier detection heuristic or any change in the training recipe, hyper-parameters, or model topology. 
Table~\ref{table:benchmark_results_v2} provides an overview of MX9 training results for a variety of 
discriminative benchmarks across different model topologies and data modalities. The results show that the models 
trained with MX9 match FP32 baseline accuracy within the error margin of run-to-run FP32-based training when 
initialized with different random seeds. The work in~\cite{nvidia2022fp8} reports FP8 results on a subset of these benchmarks.

In addition to benchmarks listed in Table~\ref{table:benchmark_results_v2}, we further evaluated the robustness of 
MX9 training on three proprietary recommendation models. Table~\ref{table:training_meta} summarizes the 
results of this evaluation. Consistent with the results of our open-sourced benchmark suite, MX9 presents a robust 
training alternative for production-scale recommendation models. In our experience, FP8-based 
training led to gradient explosion in PR-rec3 models while MX9 had a robust convergence trend.

\begin{table}[ht]
\caption{Normalized [cross] Entropy (NE) difference of MX9 training and baseline FP32 for different recommendation models. In the mixed-precision training of PR-rec3,  $>99\%$ of the compute in forward/backward passes is still in MX9. In PR-rec2 mixed-precision training, $>94\%$ of compute is kept in MX9 format. Note that although uniform MX9 training showed a robust training trend across all $3$ models (unlike FP8), for some benchmarks we opt to use a mixed-precision setting to meet the service-level agreements. 
}
  \label{table:training_meta}
  \renewcommand{\arraystretch}{1}
  \centering
  \resizebox{\columnwidth}{!}{\begin{tabular}{|l|c|c|c|}
\hline
Model Name      & Model Topology  & MX9 Training & Mixed-Precision Training \\ \hline
PR-rec1   & DLRM          &  $0.02\%$ & N/A \\ \hline
PR-rec2   & Transformer    &  $0.05\%$\ & $0.01\%$ \\ \hline
PR-rec3   & DHEN              & $0.1\%$ & $-0.02\%$ \\ \hline
\end{tabular}}
\renewcommand{\arraystretch}{1}
	\vspace{-4pt}
\end{table}

Internally, we use a tight threshold of $0.02\%$ for production models, which is set based on the training variations from run to run with the FP32 format~\cite{deng2021low}. Training PR-rec1 with MX9 and MX6 can 
both meet this requirement but MX4 results in $\sim 0.07\%$ gap with the same number of training 
iterations and hyper-parameters as the baseline FP32. MX6 and MX4 have roughly $2\times$ and $4\times$ lower circuitry compared to FP8. We leave further evaluation of MX4 and MX6 for training recommendation 
models to future work. PR-rec2 and PR-rec3 are more challenging benchmarks to quantize. As shown in 
Table~\ref{table:training_meta}, certain layers (e.g., first and last layer) should be kept in high bit-width to meet the $0.02\%$ threshold.

Table~\ref{table:generative_training} shows the result of training generative transformers of various sizes with 
MX9 format. MX9 matches the quality of FP32 without the need for any heuristic-based outlier 
detection (e.g., \texttt{transformer engine} in~\cite{nvidia2022fp8}) or change in the training recipe (number of 
training iterations, hyper-parameters, or model topology).

\begin{table}[ht]
\caption{Generative training of dense and MoE language models with MX9 matches the language model loss of the FP32 baseline.}
  \label{table:generative_training}
  \renewcommand{\arraystretch}{1}
  \centering
{\begin{tabular}{|l|c|c|}
\hline
Model                & Baseline FP32 & MX9 \\ \hline
GPT-XS (6M)         & 4.61          & 4.61 \\ \hline
GPT-S (20M)         & 4.03          & 4.03 \\ \hline
GPT-M (150M)        & 3.31          & 3.31 \\ \hline
GPT-L (345M)        & 3.11          & 3.11 \\ \hline
GPT-XL (1.5B)       & 2.74          & 2.74 \\ \hline
MoE (1.9B) & 2.22      & 2.21 \\ \hline
\end{tabular}}
\renewcommand{\arraystretch}{1}
\vspace{-4pt}
\end{table}

\begin{figure}[ht]
\includegraphics[width=0.9\columnwidth]{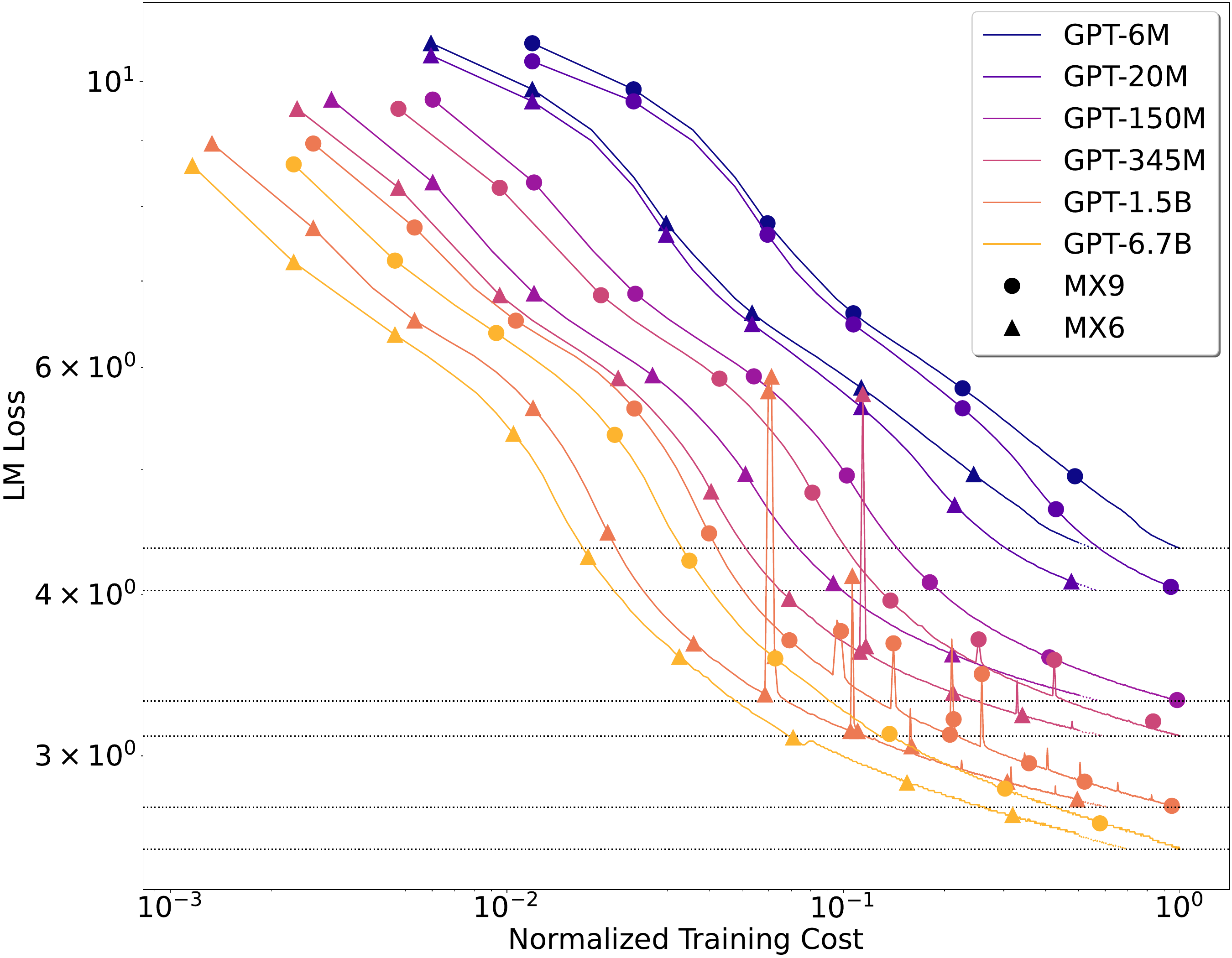}
\caption{Training generative language models with MX6 data format. The y-axis shows the LM loss of different-sized GPT models and the x-axis shows the normalized training cost with respect to a MX9 baseline. The cost is approximated based on the expected tensor unit throughput not accounting for the overhead of vector operations. Unlike MX9, training language models with MX6 requires more iterations of training to match baseline accuracy. The additional iterations are shown as dashed lines in MX6 training curves. Dotted horizontal lines show the baseline FP32 LM loss at efficiency.}
\label{fig:bdr6training}
\vspace{-16pt}
\end{figure}

Our assessment of training generative language models with the more efficient MX6 format shows that the models 
trained with this data format are also able to converge robustly. Figure~\ref{fig:bdr6training} shows the Language 
Model (LM) loss as a function of normalized cost per training iteration for MX9 versus MX6. MX6 requires more 
training iterations compared to the baseline FP32 (and by extension MX9) to converge to the same LM loss. Given 
the relative throughput of MX6, however, the model can still converge to the same quality as the baseline with an 
overall lower training cost. Quantifying the exact impact of the MX6 data format on the underlying generative 
training power laws is out of the scope of this paper.

\section{Related Work}
AI training and inferencing using narrow bit-width data formats has become a mainstream industry standard. Prior work 
in this domain can be grouped into two main categories: classic scalar floating-point and block floating-point.

The most common examples of scalar floating-point formats are BF16~\cite{intel2018bfloat,kalamkar2019bfloat}, 
TF32~\cite{nvidia2021tf32}, FP16~\cite{nvidia2017mixed}, and FP8~\cite{nvidia2022fp8, wang2018training,sun2019hybrid}. 
While the scaling down of traditional scalar floating-point formats from FP32 to FP8 has been very successful, it is 
reaching a point of diminishing returns. To continue driving the advancement of AI, there is a need for an alternative 
approach to further improve the efficiency and robustness of data encoding at narrow bit-widths for AI models.

Several prior works have proposed the use of Block Floating-Point as a promising alternative for AI training and inference~\cite{darvish2020pushing, dai2021vs, drumond2018hbfp, burcu2022accuracy, zhang2022hpca, yeh2022like}. The work in~\cite{darvish2020pushing} provides a detailed analysis of 
different block floating-point variants and their application for AI inferencing at production-scale. The work 
in~\cite{drumond2018hbfp} takes a hybrid approach for AI training using BFP. The hybrid BFP recipe is further 
optimized in~\cite{burcu2022accuracy} by varying the block size and mantissa bit-width across different layers and 
training epochs. FAST~\cite{zhang2022hpca} also varies the mantissa bit-width per training iteration and supports 
this with a bit-serial hardware architecture that multiplies two mantissa bits at a time. In another 
work~\cite{dai2021vs}, authors explored hierarchical BFP with a per-channel floating-point scale and a per-block 
integer scale to accelerate AI inference. Although promising, to the best of our knowledge, none of the prior work 
reports a generalizable basic BFP data format that can be used as a drop-in replacement for scalar floating-point, 
both in terms of model accuracy and usability (i.e., straightforward training recipes and a direct cast option for inference).

The unified quantization framework presented in this paper allows reasoning about both scalar and block floating-point 
under the same design space. Through extensive exploration, we have identified a new family of data formats that 
provide a better trade-off in terms of fidelity (model accuracy) and hardware cost. The proposed format provides a robust alternative for AI training and inferencing 
without the need for complex heuristics, or changes in the training or inferencing recipes.

\section{Conclusion}
This paper introduces BDR, a general framework for evaluating various quantization schemes. Leveraging BDR, we explore the benefit of shared microexponents in the form of MX formats. We evaluated variants of MX data formats on over 20 benchmarks across different scales, modalities, and model topologies. Our results corroborate the efficacy of BDR data formats as a promising solution for improving the performance of future AI hardware.

\section{Proof of Theorem~\ref{theorem:MSFPp_worst_case_lower_bound}} \label{Sec:proof}
We first prove Theorem~\ref{theorem:MSFPp_worst_case_lower_bound} for $N\!=\!k_{1}$ with scale $1$ and then extend it to an arbitrary $N$ and scale $s$. Consider a vector $\mathbf{X}\!=\![x_1, x_2, \cdots, x_{k_{1}}]\!\in\!\mathbb{R}^{k_{1}}$ with maximum exponent $E$ before being converted to BDR data format.
Since $E$ is the maximum shared exponent, all elements $x_{i}$ can take the binary form
\begin{equation}  \label{eq:max_exp}
\resizebox{0.6\linewidth}{!}{$\begin{aligned}
	x_{i}&= (-1)^{\psi^{(i)}} \times p_{0}^{(i)}.p_{-1}^{(i)}p_{-2}^{(i)}p_{-3}^{(i)}p_{-4}^{(i)}\cdots \times 2^{E}\\
	&=(-1)^{\psi^{(i)}} \times \big(\sum_{j=0}^{\infty} p_{-j}^{(i)} 2^{-j}\big) \times 2^{E},
	\end{aligned}$}
\end{equation}
where  $\psi^{(i)} \in \{0,1\}$, $p_{j}^{(i)} \in \{0,1\}$, and $1\leq\!i\!\leq\!k_{1}$. $E$ is the private exponent of an element $x_{\kappa}$ in $\mathbf{X}$ with the maximum absolute value among $x_i$'s.

Recall that BDR supports $m$ mantissa bits and $1$ sign bit. Since elements in $\mathbf{X}$ are in FP32 precision, which supports $8$ exponent bits, $E$ can be represented in its FP32 form. After rounding the mantissa bits to the nearest floating point number with $m$ mantissa bits, we have:
\begin{equation}  \label{eq:max_exp_round}
\resizebox{0.6\linewidth}{!}{$\begin{aligned}
	Q(x_{i})&=(-1)^{\psi^{(i)}} \times \big(\sum_{j=0}^{m-1} Q_{-j}^{(i)} 2^{-j} \big) \times 2^{E},
	\end{aligned}$}
\vspace{-0.5em}
\end{equation}
where $\psi^{(i)} \in \{0,1\}$, $Q_{-j}^{(i)} \in \{0,1\}$, and $1\leq i \leq k_{1}$. This results into the quantization error:
\begin{equation} \label{eq:q_error_max}
\begin{aligned}
\lvert  Q(x_{i}) - x_{i} \rvert \leq 2^{E-m},
\end{aligned}
\end{equation}
for all $1\leq\!i\!\leq\!k_{1}$. Similarly, the quantization error for elements with sub-block shift of $\tau \leq \beta$ is upper bounded as follows:
\begin{equation} \label{eq:q_error_max-1}
\begin{aligned}
\lvert  Q(x_i) - x_i \rvert \leq 2^{E-\tau-m}.
\end{aligned}
\end{equation}
Without loss of generality, assume that $r_{\tau}$ sub-blocks in $\mathbf{X}$ have a sub-block shift of $\tau$. This assumption implies that there exist $\sum_{\tau=0}^{\beta-1}r_{\tau} k_{2}$ elements with sub-block shifts smaller than $\beta$, since the sub-block size is $k_{2}$. The remaining $k_{1}-\sum_{\tau=0}^{\beta-1}r_{\tau} k_{2}$ elements in the block have a sub-block shift of $\beta$. Thus, according to \eqref{eq:q_error_max-1}, we have:
\begin{equation} \label{eq:q_error_all}
\resizebox{0.9\linewidth}{!}{$\begin{aligned}
	\lVert  Q(\mathbf{X}) - \mathbf{X} \rVert^{2} &= \sum_{i=1}^{k_{1}} |q(x_i)-x_i|^{2}\\
	&\leq  \sum_{\tau=0}^{\beta-1}\!r_{\tau} k_{2} (2^{E-\tau-m})^{2}\!+\!(k_{1}\!-\!\sum_{\tau=0}^{\beta-1}\!r_{\tau} k_{2}) (2^{E-\beta-m})^{2} \\
	&=(\frac{k_{1} + \sum_{\tau=0}^{\beta-1} r_{\tau} k_{2} (2^{2\beta-2\tau}-1)}{2^{2\beta}}) (2^{2E-2m}).
	\end{aligned}$}
\end{equation}
In BDR, at least one element within each sub-block with the sub-block shift of $\tau < \beta$ (not equal to $\beta$) should have a private exponent of $E-\tau$. So, there exist at least $r_{\tau}$ elements with private exponent of $E-\tau$ for $\tau < \beta$. Then, the power of signal is lower-bounded as follows:
\begin{equation} \label{eq:signal}
\resizebox{0.3\linewidth}{!}{$\begin{aligned}
	\lVert  \mathbf{x} \rVert^{2}& =\sum_{i=1}^{k_{1}}x_{i}^2 \\
	&\geq \sum_{\tau=0}^{\beta-1} r_{\tau} (2^{E-\tau})^{2} \\
	\end{aligned}$}
\end{equation}
By combining the bounds in \eqref{eq:q_error_all} and \eqref{eq:signal}, we have the following inequality for the quantization noise-to-signal ratio:
\begin{equation} \label{eq:nsr}
\resizebox{0.85\linewidth}{!}{$\begin{aligned}
	\frac{\mathbb{E}[\lVert Q(\mathbf{X}) - \mathbf{X} \rVert^{2}]}{\mathbb{E}[\lVert \mathbf{X} \rVert^{2}]} & \leq  \max_{\mathbf{X} \sim \rho} \frac{\lVert q(\mathbf{X}) - \mathbf{X} \rVert^{2}}{\lVert \mathbf{X} \rVert^{2}} \\
	&\leq \frac{\big(k_{1} + \sum_{\tau=0}^{\beta-1} r_{\tau} k_{2} (2^{2\beta-2\tau}-1)\big) (2^{2E-2m})}{2^{2\beta}\sum_{\tau=0}^{\beta-1} r_{\tau} (2^{2E-2\tau})} \\
	&= \frac{\big(k_{1} + \sum_{\tau=0}^{\beta-1} r_{\tau} k_{2} (2^{2\beta-2\tau}-1)\big) (2^{-2m})}{\sum_{\tau=0}^{\beta-1} r_{\tau} (2^{2\beta-2\tau})}\\
	&= \Big(\frac{k_{1}}{\sum_{\tau=0}^{\beta-1} r_{\tau} (2^{2\beta-2\tau})}+k_{2} \\
	& -\frac{k_{2}\sum_{\tau=0}^{\beta-1} r_{\tau} }{\sum_{\tau=0}^{\beta-1} r_{\tau} (2^{2\beta-2\tau})}\Big).2^{-2m} \\
	& \leq \big(\frac{k_{1}}{2^{2\beta}}+k_{2}-\frac{h\sum_{\tau=0}^{\beta-1} r_{\tau} }{\sum_{\tau=0}^{\beta-1} r_{\tau} (2^{2\beta-2\tau})}\big).2^{-2m}.
	\end{aligned}$}
\end{equation}
Note that the last inequality in \eqref{eq:nsr} holds since at least one sub-block containing $x_{\kappa}$ has the sub-block shift of zero; i.e., $r_0 \geq 1$. We can further use the inequality
\begin{equation} \label{eq:inequality_lemma} \nonumber
\begin{aligned}
\frac{\sum_{\tau=0}^{\beta-1} r_{\tau} }{\sum_{\tau=0}^{\beta-1} r_{\tau} (2^{2\beta-2\tau})} \geq \frac{1}{2^{2\beta}}
\end{aligned}
\end{equation}
to streamline \eqref{eq:nsr} as
\begin{equation} \label{eq:nsr_simplified}
\resizebox{0.7\linewidth}{!}{$\begin{aligned}
	\frac{\mathbb{E}[\lVert Q(\mathbf{X}) - \mathbf{X} \rVert^{2}]}{\mathbb{E}[\lVert \mathbf{X} \rVert^{2}]} & \leq  (\frac{k_{1}}{2^{2\beta}}+k_{2}-\frac{k_{2}}{2^{2\beta}}).2^{-2m} \\
	& = \big(\frac{k_{1}+(2^{2\beta}-1)k_{2}}{2^{2\beta}}\big).2^{-2m}
	\end{aligned}$}
\end{equation}
Converting \eqref{eq:nsr_simplified} into the QSNR in \eqref{eq:qsnr_def} leads to:
\begin{equation} \label{eq:snr_final}
\resizebox{0.7\linewidth}{!}{$\begin{aligned}
	\text{QSNR} 
	& \geq -10 \, \text{log} \Big( \big(\frac{k_{1}+(2^{2\beta}-1)k_{2}2}{2^{2\beta}}\big).2^{-2m}\Big) \\
	& = 10\,m \, \text{log}(4) + 10 \, \text{log} \, \big(\frac{2^{2\beta}}{k_{1}+(2^{2\beta}-1)k_{2}}\big) \\
	& \geq 6.02\,m + 10 \, \text{log} \, (\frac{2^{2\beta}}{k_{1}+(2^{2\beta}-1)k_{2}}) .
	\end{aligned}$}
\end{equation}
The lower-bound \eqref{eq:snr_final} holds for an arbitrary scale $s$, since one can change the variable $\mathbf{X}$ to $\frac{1}{s}\mathbf{X}$ and recast the QSNR as
\begin{equation} \nonumber
\resizebox{0.65\linewidth}{!}{$\begin{aligned}
	 \frac{\mathbb{E}[\lVert sQ(\frac{1}{s}\mathbf{X}) - \mathbf{X} \rVert^{2}]}{\mathbb{E}[\lVert \mathbf{X} \rVert^{2}]} = 
	 \frac{\mathbb{E}[\lVert Q(\frac{1}{s}\mathbf{X}) - \frac{1}{s}\mathbf{X} \rVert^{2}]}{\mathbb{E}[\lVert \frac{1}{s}\mathbf{X} \rVert^{2}]}.
	\end{aligned}$}
\end{equation}
 The lower-bound \eqref{eq:snr_final} also holds for $N\!>\!k_{1}$ since the noise-to-signal ratio in~\eqref{eq:nsr_simplified} is effectively averaging across blocks of size $k_{1}$. For $N\!<\!k_{1}$, we are effectively reducing the BDR block size to $N$ which improves the lower-bound if we replace $k_{1}$ with $N$. In summary, the following lower-bound holds for various values of $N$ and scale $s$:
\begin{equation} \nonumber
\resizebox{0.8\linewidth}{!}{$\begin{aligned}
\text{QSNR} \geq 6.02\,m + 10 \, \text{log} \, (\frac{2^{2\beta}}{(2^{2\beta}-1)k_{2}+\text{min}(N,k_{1})}). \hspace{0.5cm} \blacksquare
\end{aligned}$}
\end{equation}

\newpage
\bibliographystyle{IEEEtranS}
\bibliography{refs}


\end{document}